# 다중 신경망 레이어에서 특징점을 선택하기 위한 전이 학습 기반의 AdaBoost 기법

주마벡*, 가명현*, 고승현*, 조근식*
*인하대학교 컴퓨터정보공학과
e-mail : jumabek4044@gmail.com

## Transfer Learning based on Adaboost for Feature Selection from Multiple ConvNet Layer Features

Jumabek Alikhanov*, Myeong Hyeon Ga*, Seunghyun Ko*, Geun-Sik Jo*
*Dept. of Computer Science and Information Engineering, Inha University

## Abstract

Convolutional Networks (ConvNets) are powerful models that learn hierarchies of visual features, which could also be used to obtain image representations for transfer learning. The basic pipeline for transfer learning is to first train a ConvNet on a large dataset (source task) and then use feed-forward units activation of the trained ConvNet as image representation for smaller datasets (target task). Our key contribution is to demonstrate superior performance of multiple ConvNet layer features over single ConvNet layer features. Combining multiple ConvNet layer features will result in more complex feature space with some features being repetitive. This requires some form of feature selection. We use AdaBoost with single stumps to implicitly select only distinct features that are useful towards classification from concatenated ConvNet features. Experimental results show that using multiple ConvNet layer activation features instead of single ConvNet layer features consistently will produce superior performance. Improvements becomes significant as we increase the distance between source task and the target task.

## 1. Introduction

Though, originally Neocognitron [1] introduced back in 80s, due to lack of large scale dataset and computational power, ConvNets were not widely used in both academia and industry until Krizhevsky [2]'s breakthrough on ILSVRC-12 [3] visual recognition challenge.

Since success of deep learning methods depend on large scale dataset and heavy computation it would be a good idea to employ a trained ConvNet to other tasks where small datasets or lack of equipment make it impossible or expensive to train specific network. Previous works [4], [5], [6], [7], [8], [9] demonstrated transfer learning by using certain layer activation features of ConvNet as an image representation for their target task. They extract ConvNet features from certain layer, for instance, fully connected layer 6 of ConvNet (from now on we will use the notation FC6) and train linear SVM on FC6 ConvNet features for their classification tasks. In this work we will study if using multiple ConvNet layer activation features instead of single ConvNet layer features will produce better results. Unlike other works we are proposing to use multiple ConvNet layer features as we explain the motivation later in paper (in section 3.1). However, when we combine those multiple layer features our feature space becomes much more complex than single ConvNet layer features that used in previous works. Hence, we need feature selection in order to simplify our feature space. We select features implicitly by using AdaBoost with single stump weak classifier. For example, we first concatenate all the fully connected layer features of FC6, FC7, FC8 then train Adaboost with single stump on those features. Since single decision stumps consider only one feature at a time, Adaboost will implicitly select the only distinct, helpful features from among the concatenated ones. We will cover this in detail in section 3.

## 2. Related Work

In [8] authors with their visualization technique analyzed and improved ConvNet architecture of Krizhevsky [2] and won the ILSVRC-13. They also demonstrated how their ConvNet, trained on ImageNet, generalizes well to the Caltech-256 dataset. They trained linear SVM on ConvNet layer features and showed that activation features extracted from the later layers of ConvNet produces robust performance when used as a descriptor for the target task. By following the same pipeline, other research [4], [6], [7], [9] employed the ConvNet layer activation features with linear SVM. In [7] authors conducted experiments on a series of visual recognition tasks. The experiments consistently produced superior results, compared to state-of-the-art, highly tuned methods that use conditional handcrafted features like SIFT, HOG and LBP. They demonstrated that simple augmentation techniques, such as jittering, boost performance significantly. In all of these works linear SVM with single ConvNet layer features is employed. However, in our work we intend to improve performance by combining multiple ConvNet layers. In other words, we concatenate multiple ConvNet layer features in order to obtain better image representation. Since we concatenate multiple ConvNet layer features, resulting feature space will become more complex as many features are repetitive. This makes some form of feature selection necessary. We use AdaBoost with single stump weak learner to tackle the problem of feature selection.

## 3. AdaBoost with Single Stumps as Feature Selector Classifier

In this section we will give the motivation for why one has to consider combining multiple ConvNet layers and rationale for choosing AdaBoost as a classifier.

### 3.1 ConvNet Layer Descriptions in Transfer Learning

Different ConvNet layers address the various aspects of signal. There is a difference of layers when describing images in transfer learning. This characteristic surely affects the performance of transfer learning. To illustrate the notion of difference of ConvNet layers in describing images, consider a simple example. Assume we have dataset D, with five test images in its test set (Dtest={"cat", "dog", "horse", "sheep", "camel"}), which we want to classify. Suppose we train two classifiers for dataset D: i) classifier "C6" on FC6 features (remember that FC6 stands for activations of the fully connected layer 6 of the ConvNet), ii) classifier "C7" on FC7 features. Assume that "C6" predicts "cat", "dog", "horse" test images correctly while "C7" gives correct predictions for "cat", "horse" and "sheep" test images. From the prediction results, we can conclude that FC6 features are good at describing an image of "dogs" whereas FC7 features are good at describing images of "sheep". Now, if we were to combine these two FC6 and FC7 ConvNet features in some 'clever way', we would get the correct classification for the test images "cat", "dog", "horse", and "sheep". In this way, a combined multiple ConvNet layer features would perform better (4 out of 5) than single ConvNet layer activation features (3 out of 5). An insight one can obtain from this example is that we have to take into account ConvNet layers' difference in describing images when we do transfer learning in order to fully transfer ConvNet learning. Once we know that next step is to combine those ConvNet layer features so that we can transfer learning of the ConvNet in a more complete fashion comparing to learning that would be transferred with single ConvNet layer features

### 3.2 Feature Selection

While the behavior of ConvNet layers differs (see section 3.1), that does not mean all of the 4096-dimensional features of FC6 encode distinct characteristics of an image with respect to the other 4096-dimensional features of FC7. Rather, small number of features from FC6 and FC7 together would form a complementary features. For instance, let us say only 3000 features out of 8192 are useful and distinct. This means when we concatenate those two fc6 and fc7 ConvNet layers to get the final 8192-dimensional feature descriptor we will get a somewhat better image description compared to using single ConvNet layer features. But most of the features will be repetitive and not helpful towards describing an image. As a matter of fact some features might even become a noise. Since not all of the combined features are helpful for describing the image, we face the problem of selecting only those distinct, helpful features. In other words, we need to find some method that will select only the features from among 8192-dimensional features that are helpful for classification. For both to train a classifier and to select only helpful features, we use AdaBoost with single decision stumps being a weak classifier. Decision stumps select the best feature that will decrease the loss with respect to the current weights that calculated by AdaBoost. In other words in each step decision stump will select the best feature and a threshold that will give the highest classification accuracy. This way, AdaBoost with single decision stumps will implicitly select only those distinct and helpful features from among the 8192-dimensional features. Hence, we use AdaBoost with single decision stumps weak learner to take advantage of multiple ConvNet layer features. Note that, for clarity we showed a situation when we concatenate only two FC6 and FC7 layer features, however in our experiments we use three fully connected layer features as our image representation.

## 4. Results

Our experiments intend to compare the performance of multiple ConvNet layer features against single ConvNet layer features. As covered in section 3 AdaBoost with single stumps will implicitly select the only 'good' features among combined ConvNet layer features and also from single ConvNet layer features. Table 1 represents the experimental results for three standard classification datasets. First three rows corresponds to classification accuracy when single ConvNet layer features are used. Last row displays the results for concatenation of three fully connected layers of the ConvNet. In Both cases AdaBoost with single stumps is trained as a classifier and implicit feature selector. Datasets are ordered from left to right based on the distance between them and the source task (ILSVRC-12) dataset. Among our datasets Caltech-256 [10] is the closest to target task dataset which is ILSVRC-12 in our case (see section 4.2). Because it contains 256 common real world objects where many of them exists in ILSVRC-12 as well. SUN397 [12] on the other hand, furthest dataset among those three. Because it contains scenes not the objects. The pattern one can notice from Table 1 is that as the distance between source task and target task grows improvement in performance that we achieve from multiple ConvNet layer features increases.

### 4.1 Datasets
Experiments are conducted on Caltech-256 [10], VOC07 [11] and SUN397 [12] standard classification datasets.
1) *Caltech-256*: Contains around 30K image for 257 categories, including a cluttered category. Each category consists of at least 100 images. Following the lead of [13], we split the dataset by taking 60 images from each class for training set, and the rest for the test set.

2) *VOC07*: VOC07 contains 5011 images in its training and validation set together, and 4952 images in the test set. We used training and validation sets as our training set as was done elsewhere [7], [9].
3) *SUN397*: One of the challenging datasets for scene classification. It contains 108K images of 397 classes with at least 100 images in each category. We took 50 images for both training and test subsets from each category, as was done in [12].

**4.2 Transfer Learning**

In ILSVRC-12 Kirizhevsky won the first place. His trained network called AlexNet is now publicly available. If we use knowledge of AlexNet for other task, say for VOC07 object classification dataset. We would be transferring the learning from ILSVRC-12 (source task) to VOC07 (target task). To conduct our transfer learning experiments, we use AlexNet's activation features as our image representation. To obtain those, we give an image as in input to a ConvNet (AlexNet) and extract activation features from certain fully connected layers. To carry out feed forward computations we use Caffe [13] software.

|  | Caltech-256 | VOC07 | SUN397 |
|---|---|---|---|
| FC6 | 69.5 | 67.4 | 47.3 |
| FC7 | 72.5 | 70 | 47.4 |
| FC8 | 70.8 | 69.1 | 42.9 |
| FC6-FC7-FC8 | **72.8** | **71** | **49.1** |

**Table 1: Classification results for three datasets (%).** FC6, FC7, FC8 notation stands for fully connected layer 6, 7, 8 of ConvNet (AlexNet in our case) respectively. FC6-FC7-FC8 stands for combination of three fully connected layers.

**5. Conclusion**

In this paper we demonstrated multiple ConvNet Layer features performance being superior to single ConvNet layer features. We used AdaBoost with single stumps as it selects the distinct, helpful features implicitly. From the experimental results we can conclude that indeed multiple ConvNet layer features perform better. Feature space complexity could be solved by using some form of feature selection. In our case we used AdaBoost with single stumps as a feature selector.

**Acknowledgement**

We gratefully acknowledge the support of NVIDIA Corporation with the donation of the Tesla K40 GPU to this research.

This work was supported by the National Research Foundation of Korea(NRF) grant funded by the Korea government(MSIP) (No. 2015-R1A2A2A03006190).